\pdfoutput=1

\documentclass[11pt]{article}

\usepackage[]{acl}

\usepackage{comment}
\usepackage{rotating}
\usepackage[utf8]{inputenc}
\usepackage{times}
\usepackage{latexsym}
\usepackage{rotating,tabularx,siunitx}

\usepackage{float}
\usepackage{graphicx}
\usepackage{subfigure}
\usepackage{caption}
\usepackage{amsmath,amssymb}
\usepackage{mathtools}
\usepackage{booktabs}
\usepackage{multirow}
\usepackage{siunitx}
\usepackage{adjustbox}
\usepackage{pifont}

\usepackage{tabularray}

\usepackage[T1]{fontenc}

\usepackage[utf8]{inputenc}

\usepackage{microtype}
\usepackage{amsmath}
\usepackage{amsfonts}
\usepackage{multirow}

\usepackage{tabularx}
\usepackage{longtable}

\usepackage[ruled]{algorithm2e}
\usepackage{algpseudocode}
\usepackage{hyperref}
\usepackage[capitalise]{cleveref}
\usepackage{stfloats}
\usepackage{color}
\usepackage{xcolor}
\definecolor{my_red}{rgb}{0.98039216, 0.49803922, 0.43529412}
\definecolor{my_green}{rgb}{0.56078431, 0.81568627, 0.78823529}

\usepackage{graphicx}
\usepackage{subfigure}
\usepackage{longtable}
\usepackage{makecell}
\usepackage{enumitem}
\usepackage{amssymb}
\usepackage{multirow}
\usepackage{svg} 
\usepackage{diagbox}

\newtheorem{definition}{Definition}
\usepackage{bm}
\usepackage{hyperref}
\usepackage{defination}

\title{SecFormer: Fast and Accurate Privacy-Preserving Inference for Transformer Models via SMPC}

\author{{\bf Jinglong Luo}$^{1,2}$\quad{\bf Yehong Zhang}$^{2,}$\thanks{Corresponding author}\quad{\bf Zhuo Zhang}$^{1,2}$\quad{\bf Jiaqi Zhang}$^{2}$\quad{\bf Xin Mu}$^{2}$ \\{\bf Hui Wang}$^{2}$\quad{\bf Yue Yu}$^{2}$\quad{\bf Zenglin Xu}$^{1,2,*}$\\
$^{1}$Harbin Institute of Technology, Shenzhen 
$^{2}$Peng Cheng Laboratory, Shenzhen, China \\
\texttt{\{luojl, zhangyh02, zhangjq02, mux, wangh06, yuy\}@pcl.ac.cn},\\
 \texttt{\{iezhuo17, zenglin\}@gmail.com}
\\
}

\begin{document}
\maketitle

\begin{abstract}
With the growing use of Transformer models hosted on cloud platforms to offer inference services, privacy concerns are escalating, especially concerning sensitive data like investment plans and bank account details. Secure Multi-Party Computing (SMPC) emerges as a promising solution to protect the privacy of inference data and model parameters. However, the application of SMPC in Privacy-Preserving Inference (PPI) for Transformer models often leads to considerable slowdowns or declines in performance.
This is largely due to the multitude of nonlinear operations in the Transformer architecture, which are not well-suited to SMPC and difficult to circumvent or optimize effectively.
To address this concern, we introduce a comprehensive PPI framework called \textit{SecFormer} to achieve fast and accurate PPI for Transformer models. We successfully eliminate the high-cost exponential and maximum operations in PPI without sacrificing model performance and develop a suite of efficient SMPC protocols by employing suitable numerical computation methods to boost other complex nonlinear functions in PPI, including GeLU, LayerNorm, and a redesigned Softmax. Our extensive experiments reveal that \textit{SecFormer} outperforms \textit{MPCFormer} in performance, showing improvements of $3.4\%$ and $24.7\%$  for BERT$_{\text{BASE}}$ and BERT$_{\text{LARGE}}$, respectively. In terms of efficiency, \textit{SecFormer} is 3.57 and 3.58 times faster than \textit{PUMA} for BERT$_{\text{BASE}}$ and BERT$_{\text{LARGE}}$, demonstrating its effectiveness and speed. The code is available by clicking \href{https://github.com/jinglong696/SecFormer}{here}.
\end{abstract}
\section{Introduction}
Transformer models~\citep{Vaswani-2017-Attention, bert, gpt2, gpt3, 2020t5, liu2019roberta, lewis2020bart, zeng2021pangu, instructGPT, OpenAI2023GPT4TR} demonstrate exceptional performance across diverse downstream tasks~\cite{chan-etal-2024-exploring, jiayang-etal-2023-storyanalogy,  LuZXLZX19,ZhangLLLBBX20,LuR0LX20,LiuYZLLBX20, zhang2023fedlegal,WangWQFXNWKFL23} and are extensively employed in a Model-as-a-Service (MaaS) paradigm to deliver high-quality inference services to clients. However, this MaaS framework poses a significant privacy risk~\cite{li2023privacy, li-etal-2023-multi-step, li-etal-2023-sentence, li2024privlmbench, zhang2024revisiting, zhang2020additively} for inference data and model parameters. For instance, both Copilot\footnote{https://github.com/features/copilot} and ChatGPT\footnote{https://chat.openai.com}, which are Transformer-based services, necessitate users to upload plaintext requests. This operational procedure not only poses a threat to users' privacy but also probably contravenes relevant legal regulations such as the EU’s General Data Protection Regulation
(GDPR)\footnote{https://gdpr-info.eu/}.

\begin{figure}[t]
\centering
\label{fig:example_figure}
\hspace{-2.9mm}\subfigure{
    \label{fig:time_breakdown}
    \includegraphics[scale=0.25]{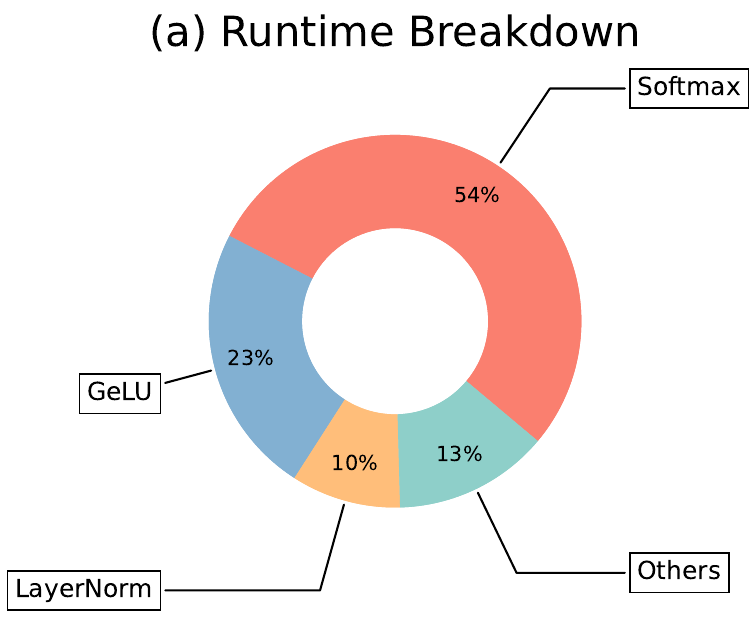}
}
\hspace{-2.29mm}\subfigure{
    \label{fig:performance_influence}
    \includegraphics[scale=0.25]{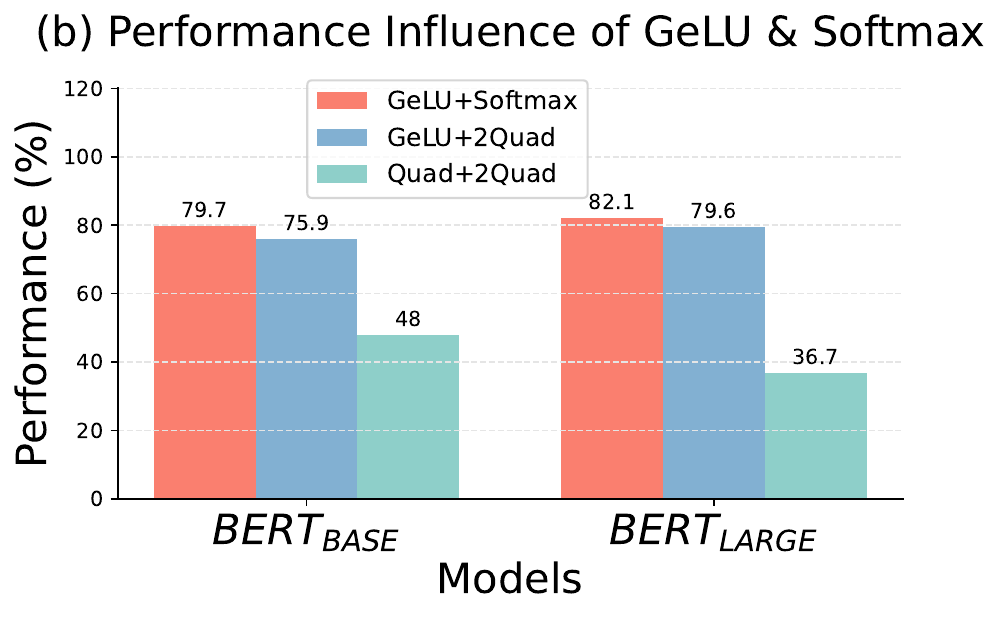}
}
\caption{(a) Runtime breakdown of the BERT$_{\text{BASE}}$ model (12 layers, 512 tokens) based on an SMPC library. The total runtime for an example is 71 seconds. (b) Influence of different activation functions on model performance.}
\end{figure}
 
Secure Multi-Party Computation (SMPC) \citep{shamir1979share, Yao1986MPC, GoldreichMW87}, has demonstrated great potential in safeguarding the privacy of both inference data and model weights \citep{gilad2016cryptonets, liu2017oblivious, mishra2020delphi, rathee2021sirnn, HuangLHD22, luo2023practical,zhang2023survey, zhang2024survey}. However, conducting Privacy-Preserving Inference (PPI)\footnote{Without confusion, we refer to SMPC-based PPI as PPI for short in this paper .} for Transformer models using SMPC proves to be notably slow. To illustrate, BERT$_{\text{BASE}}$ \citep{bert} takes $71$ seconds to inference per sample via SMPC, while plain-text inference takes less than $1$ second. 

This inefficiency stems from the limitations of current SMPC protocols in executing nonlinear operations in Transformer models (\cref{sec:background-2}). As depicted in \cref{fig:time_breakdown}, we find that Softmax and GeLU, which account for a small share of the plaintext inference overhead, take up $77.03\%$ of the time in PPI. This observation aligns with findings in \citet{wang2022characterization, li2022mpcformer}. In an effort to mitigate this, \citet{li2022mpcformer} \emph{redesigned the Transformer model} by substituting Softmax and GeLU with some SMPC friendly quadratics, bypassing the privacy-preserving computations of the non-linear operations (i.e., erf, exponential, and maximum) in Softmax and GeLU. This aggressive substitution significantly enhances PPI efficiency, but unfortunately, substantially diminishes the model's performance and is not scalable to larger models (\cref{fig:performance_influence}). Some other studies~\citep{dong2023puma} tried to boost the PPI by \emph{designing more efficient SMPC protocols}, which can preserve the model performance but still face expensive PPI overhead. 

In this study, we present a comprehensive PPI framework named SecFormer, tailed to achieve fast and accurate PPI for Transformer models by exploiting the superiorities of both Transformer model and SMPC protocol designs. \textit{Our investigation reveals that preserving accurate computation of GeLU significantly improves PPI performance (\cref{fig:performance_influence})}. Building on this insight, SecFormer implements model design to bypass the expensive nonlinear PPI operations such as exponential and maximum in Softmax~(\cref{sec: Overview}). This adaptation, coupled with the strategic use of knowledge distillation, allows SecFormer to construct a Transformer model that is both high-performing and compatible with SMPC. To further enhance the PPI performance,
we turn to protocol design and develop a suite of efficient SMPC protocols that utilize suitable numerical calculation methods to handle other complex nonlinear functions in PPI, such as GeLU, LayerNorm, and the redesigned Softmax~(\cref{sec: algorithms}). To be specific, SecFormer introduces a novel SMPC protocol for GeLU based on segmented polynomials and Fourier series, facilitating efficient and accurate computation of GeLU. In addition, SecFormer deploys efficient privacy-preserving square-root inverse and division calculation for LayerNorm and Softmax using the Goldschmidt method \citep{goldschmidt1, goldschmidt2}, coupled with input deflation techniques to bypass the nonlinear initial-value computation.


We conducted extensive evaluations of SecFormer on various datasets using Transformer models BERT$_{\text{BASE}}$ and BERT$_{\text{LARGE}}$. The experimental results reveal that SecFormer achieves an average performance improvement of $3.4\%$ and $24.7\%$ for BERT$_{\text{BASE}}$ and BERT$_{\text{LARGE}}$, respectively, compared to the state-of-the-art approach based on pure model design (\cref{sec: performance}), while maintaining comparable efficiency. In comparison to the approach that only improves the SMPC protocols, SecFormer exhibits a speedup of $3.57$ and $3.58$ times in PPI (\cref{sec: efficiency}), while sustaining comparable PPI performance.

\section{Background and Related Works}\label{sec:background}
\subsection{Workflow of SMPC-based Model Inference}
Secure Multi-Party Computation (SMPC) is a cryptographic technique that offers a promising solution for model Privacy-Preserving Inference (PPI) among multiple participants \citep{gilad2016cryptonets, liu2017oblivious, mishra2020delphi, rathee2021sirnn, HuangLHD22}. 
Typically, participants adhere to cryptographic primitives like secret sharing \citep{shamir1979share, GoldreichMW87} to safeguard the model weights and inference data. 
This paper mainly introduces the 2-out-of-2 secret sharing scheme due to its efficiency and representativeness. Specifically, the 2-out-of-2 secret sharing divides a secret $x$ in the ring of integers $\mathcal{Z}_L$ into two random shares $[\! [x]\! ] = ([x]_0, [x]_1)$ with $x = (([x]_0 + [x]_1) \mod L$), ensuring that neither share reveals information about $x$ while allowing correct reconstruction of $x$ when the two shares are combined. In constructing the SMPC protocols, the shares are owned by two distinct participants. They communicate the masked intermediate results to each other to accomplish the privacy-preserving computation of different functions and get the shares of the computational results.

\begin{figure}[t]
	\centering
\includegraphics[width=0.48\textwidth]{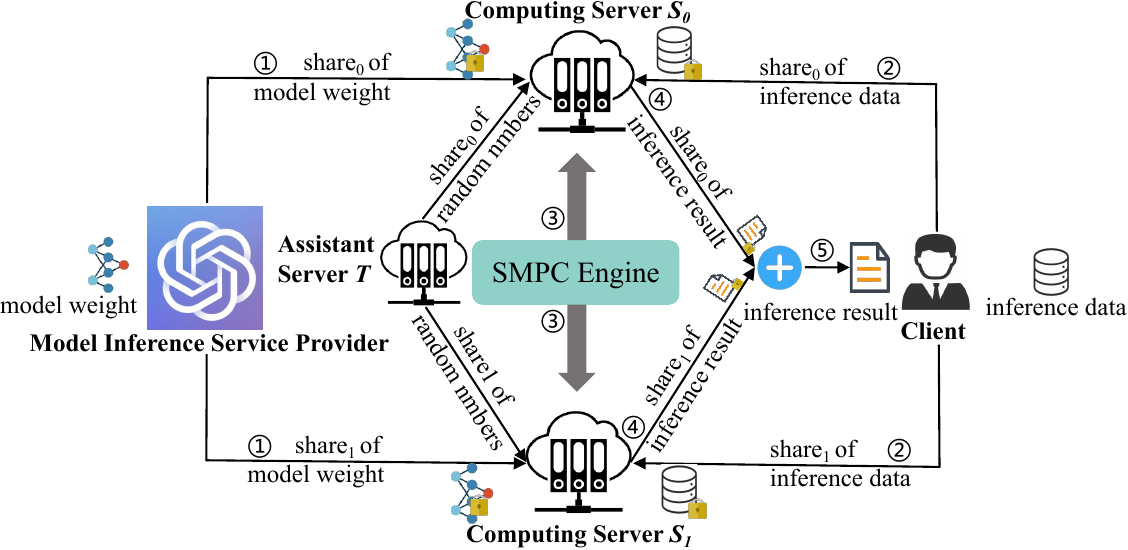} 
	\caption{Workflow of PPI based on secret sharing.}
	\label{fig: workflow}
 \vspace{-3mm}
\end{figure}

The PPI workflow leveraging 2-out-of-2 secret sharing is depicted in \cref{fig: workflow}. It involves three essential stakeholders: a \textit{model inference service provider} that needs to protect model weights, a \textit{client} that needs to protect inference data, and an \textit{SMPC engine} that performs model PPI. 
The SMPC engine contains three non-colluding servers (i.e., participants): two computing servers $S_j$ for $j \in \{0,1\}$ for shares computation of PPI and an assistant server $T$ for generating random numbers needed to execute the SMPC protocols. 
Initially, the service provider and client securely transmit the shares of model weights and inference data to $S_0$ and $S_1$, respectively (\textcircled{1} and \textcircled{2}). Subsequently, the computing servers utilize these shares as input and complete PPI by executing the SMPC protocols through interactive computation with the assistance of $T$, yielding the shares of the inference results. (\textcircled{3}).
These shares are then relayed to the client (\textcircled{4}), facilitating the local reconstruction of the inference result (\textcircled{5}). 
Since each participant has only one share of the inputs, outputs, or intermediate results, this PPI workflow can guarantee the privacy of model weights and inference data.

\subsection{Main Bottlenecks of SMPC-based Transformer Model Inference}\label{sec:background-2}
Although the above PPI workflow guarantees the privacy of model weights and inference data, it faces unacceptable communication overhead (\cref{tab: smpc protocols}) in implementing some of the nonlinear operations (i.e., Softmax, GeLU, and LayerNorm), which are abundantly present in Transformer models and become a main bottleneck in PPI.


%

Specifically, for a vector $\bm{x} = (x_1, x_2, \dots,x_{n})$, Softmax in Transformer converts it to an $n$-dimensional probability distribution with
\begin{equation}\label{eq:Softmax1}
    \text{Softmax}(\bm{x})[i] = \frac{e^{x_i-\tau}}{ \sum^{n}_{h=1}e^{x_h-\tau}}, 
\end{equation}
where $\tau = \max_{h=1}^n x_h$ is used to ensure stable numerical computations.
As indicated in \cref{tab: smpc protocols}, there are three obstacles to the SMPC of Softmax: \textit{exponential, division, and maximum}. Note that the calculation of maximum needs to call $\Pi_{LT}$ operation $\log^n$ times \citep{knott2021crypten} and becomes the biggest obstacle. 

\begin{table}[ht]
    \centering
\resizebox{\linewidth}{!}{
    \begin{tabular}{l|c|c|c|r}
    \toprule
          Notation & Input & Output & Comm Round & Comm Volume (bit)\\
         \hline
         $\Pi_{Add}$ &$([\! [x]\! ], [\! [y]\! ])$  & $[\! [x + y]\! ]$ & $0$ & $0$\\
         $\Pi_{Sin}$  &$[\! [x]\! ] $   & $[\! [sin(x)]\! ]$   & $1$ & $42$\\
         $\Pi_{Square}$ & $[\! [x]\! ]$  &  $[\! [x^2]\! ]$ & $1$ & $128$\\
         $\Pi_{Mul}$ & $([\! [x]\! ], [\! [y]\! ])$ & $[\! [xy]\! ] $& $1$ &$256$\\
         $\Pi_{MatMul}$ & $([\! [X]\! ], [\! [Y]\! ])$ & $[\! [XY]\! ] $& $1$ &$256n^2$\\
        $\Pi_{LT}$ & $([\! [x]\! ],c)$  & $[\! [(x < c)]\! ] $ & $7$ &$3456$\\
        $\Pi_{Exp}$  &$[\! [x]\! ] $   & $[\! [e^x]\! ]$  & $8$ &$1024$\\
        $\Pi_{rSqrt}$ & $[\! [x]\! ]$  & $[\! [\sqrt{x}]\! ] $& $9 + 3t$ & $6400$\\  
        $\Pi_{Div}$  &$[\! [x]\! ] $   & $[\! [1/x]\! ]$  & $16 + 2t$ & $10368$\\
    \bottomrule
    \end{tabular}}
    \caption{SMPC protocols from \citet{knott2021crypten, zheng2023secure}. $t$ is the number of Newton iterations for implementing the protocol; $n$ is the dimension of the matrix. These protocols are invoked in a black-box manner in this paper. The details are provided in \cref{sec: underlying protocols}.}
    \label{tab: smpc protocols}
    \vspace{-3mm}
\end{table}

The function of GeLU is defined as
\begin{equation}\label{eq:gelu}
    \text{GeLU}(x) = \frac{x}{2}\big (1+\text{erf}(\frac{x}{\sqrt{2}})\big), 
\end{equation}
where $\text{erf}(x) = \frac{2}{\sqrt{\pi}}\int_{0}^{x}e^{-t^2}dt$.
The GeLU function's nonlinear component is derived from the erf and \emph{there is currently no SMPC protocol for its privacy-preserving computation}.

Given a vector $\bm{x} = (x_1, x_2, \dots,x_{n})$, the LayerNorm function is defined as
\begin{equation} \label{eq:norm}
    \text{LayerNorm}(\bm{x}) = \gamma \cdot \frac{\bm{x} - \bar{x}} {\sqrt{var(\bm{x})+ \epsilon}} + \beta, 
\end{equation}
where $\bar{x} = \sum^{n}_{h = 1} x_h /n, var(\bm{x}) = \sum^{n}_{h = 1} (x_h - \bar{x})^2$, $\gamma$ and $\beta$ are two learnable parameters, and $\epsilon$ is a very small decimal used to prevent the denominator from being zero. For SMPC, the main bottleneck in computing LayerNorm comes from the \emph{division and square root operations}. 


\subsection{Efficient PPI for Transformer Models}
To alleviate the aforementioned bottlenecks, existing works on PPI for Transformer models improve the efficiency through either model design or SMPC protocol design. The studies based on model design \cite{chen2022x, li2022mpcformer, zeng2022mpcvit, SAL-VIT, liang2023merge} bypass the high overhead operations in PPI by replacing the SMPC-unfriendly nonlinear operations in Transformer. These schemes substantially increase efficiency but usually lead to a significant degradation in model performance.
The studies that design more efficient SMPC protocols \cite{hao2022iron, zheng2023primer, sigma, dong2023puma, chipherGPT, ding2023east, BLOT} improve the efficiency of PPI by customizing efficient SMPC protocols for the nonlinear operators in the Transformer. These schemes preserve the Transformer model's performance but still face expensive computational and communication overheads.

As a representative work based on model design, \citet{li2022mpcformer} improves the efficiency of PPI by replacing GeLU and Softmax with $\text{Quad} = 0.125x^2 + 0.25x + 0.5$ and 
\begin{equation}\label{eq: 2quad}
    \text{2Quad}(\bm{x})[i] = \frac{(x_i + c)^2}{\sum^{n}_{h=1}(x_h + c)^2}, 
\end{equation}
respectively, such that the privacy-preserving computation of erf, exponential, and maximum is bypassed. Following this, knowledge distillation is employed, with the fine-tuned Transformer model acting as the teacher and the approximate Transformer model as the student. Distillation is carried out on downstream task data, yielding a Transformer model compatible with SMPC. This approach is effective in improving the efficiency of PPI, however, it leads to a significant decrease in model performance. Our investigation reveals that preserving accurate computation of GeLU significantly improves PPI performance. \citet{dong2023puma} achieves the first SMPC protocol for GeLU functions by utilizing segmented functions and polynomial fitting. However, the computation of segmented functions and polynomials requires multiple calls of $\Pi_{LT}$ and $\Pi_{Mul}$, making it inefficient.

\section{SecFormer Framework}\label{sec: method}

As discussed above, existing efficient PPI studies suffer from either performance degradation or high inference overhead. To resolve this issue, 
the SecFormer framework is proposed in this section. We begin with an overview of SecFormer in Section \ref{sec: Overview} and introduce the new efficient SMPC protocols for GeLU, LayerNorm, and Softmax in Section \ref{sec: algorithms}.

\begin{figure*}[th]
    \centering
    \includegraphics[width=\textwidth]{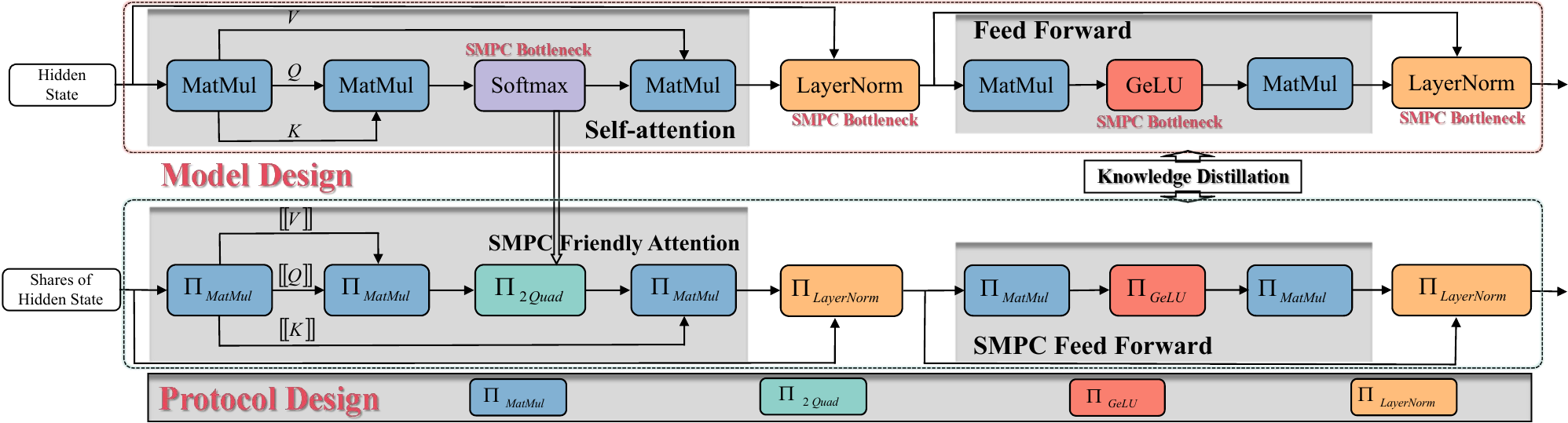}
    \caption{\textbf{An illustration of our proposed SecFormer framework.} In the model design phase, SecFormer substitutes Softmax with 2Quad to obtain an SMPC-friendly model while preserving model performance. In the SMPC protocol design stage, SecFormer improves the efficiency of the main bottlenecks in PPI for Transformer models, i.e., GeLU, LayerNorm, and 2Quad.}
    \label{fig:overall of SecFormer}
\end{figure*}

\subsection{Overview}\label{sec: Overview}
SecFormer enhances the efficiency of PPI for Transformer models, addressing both model and SMPC protocol design. The overall depiction of SecFormer is presented in~\cref{fig:overall of SecFormer}.

In the model design phase, SecFormer implements new strategies to bypass the nonlinear operations with the high overhead in PPI, such as exponential and maximum in Softmax, while preserving model performance. Specifically, SecFormer replaces Softmax with 2Quad while retaining the accurate computation of the GeLU. Inspired by \citep{li2022mpcformer}, SecFormer further improves the performance of PPI inference by incorporating knowledge distillation techniques.


In the SMPC protocol design phase, SecFormer introduces a suite of efficient SMPC protocols by utilizing appropriate numerical computation methods. Specifically, SecFormer introduces a novel SMPC protocol for GeLU based on segmented polynomials and Fourier series, which facilitates the efficient and accurate computation of GeLU. Subsequently, SecFormer deploys streamlined privacy-preserving calculation for square-root inverse and division using the Goldschmidt method \citep{goldschmidt1, goldschmidt2}, coupled with input deflation techniques to eliminate the need for nonlinear initial-value computation.

\begin{algorithm}\footnotesize
    \caption{Privacy-preserving GeLU}
    \label{alg:pp-gelu}
    \LinesNumbered
    \SetNoFillComment
    \DontPrintSemicolon
    \KwIn{For $j \in \{0, 1\}$, $S_j$ holds the shares $[x]_j$.}
    \KwOut{For $j \in \{0, 1\}$, $S_j$ returns the shares $[y]_j$, where $y = \text{GeLU}(x)$.}
    \BlankLine
    \tcc{Determine the input interval}
$[\![c_0]\!] = \Pi_{LT}([\![x]\!], -1.7)$ \quad\tcp{$(x < -1.7)$}
$[\![c_1]\!] = \Pi_{LT}([\![x]\!], 1.7)$ \quad~~~\tcp{$(x < 1.7)$}
$[\![z_0]\!] = [\![c_0]\!]$, \quad\quad\quad\quad\quad~~~\tcp{$(x < -1.7)$}
$[\![z_1]\!] = [\![z_1]\!] - [\![z_0]\!]$, \quad\quad~~~\tcp{$(-1.7 \leq x \leq 1.7)$}
$[\![z_2]\!] = 1 - [\![z_1]\!]$, \quad\quad\quad~~~ \tcp{$(x > 1.7)$}
\tcc{Compute $f(\frac{x}{\sqrt{2}})$}
$[\![\hat{x}]\!] = \frac{1}{\sqrt{2}}[\![\hat{x}]\!]$\;
$[\![\hat{x}^2]\!] = \Pi_{Square}([\![\hat{x}]\!])$\; 
$[\![\hat{x}^3]\!] = \Pi_{Mul}([\![\hat{x}]\!],[\![\hat{x}^2]\!])$\; 
$[\![\hat{x}^5]\!] = \Pi_{Mul}([\![\hat{x}^2]\!],[\![\hat{x}^3]\!])$\; 
$[\![\hat{x}^7]\!] = \Pi_{Mul}([\![\hat{x}^2]\!],[\![\hat{x}^5]\!])$\;
$[\![f(\hat{x})]\!] = -0.0031673043[\![\hat{x}^7]\!] + 0.0493278356 [\![\hat{x}^5]\!] -0.297453931[\![\hat{x}^3]\!] +  1.09952043[\![\hat{x}]\!]$\;
    \tcc{Compute $\text{erf}(x)$}
$[\![\text{erf}(x)]\!] = [\![z_0]\!] + \Pi_{Mul}([\![z_1]\!], [\![f(\hat{x})]\!]) + [\![z_2]\!]$\;
    \tcc{Compute $\text{GeLU}(x)$}
    $[\![y^\prime]\!] =  1 + [\![\text{erf}(x)]\!]$\;
 $[\![y]\!] = \Pi_{Mul}([\![\frac{x}{2}]\!], [\![y^\prime]\!])$
\end{algorithm}

\subsection{SMPC Protocols of SecFormer}\label{sec: algorithms}
We next present new efficient SMPC protocols of GeLU, LayerNorm, and the approximated Softmax designed in SecFormer.
These algorithms' security proofs and communication complexity analysis are presented in \cref{sec: security proof & comm complexity}.

\paragraph{Protocol for GeLU.}\label{sec:pp-gelu}
To address the efficiency challenges of GeLU private computations (Section~\ref{sec:background-2}), some studies replaced GeLU in \eqref{eq:gelu} with its SMPC-friendly alternatives such as ReLU~\citep{zeng2022mpcvit} or  quadratics~\citep{li2022mpcformer}. Although this approach can enhance PPI efficiency, it may result in irreversible performance losses. \citet{dong2023puma} introduces the first SMPC protocol for GeLU using segmented functions and polynomial fitting whose computation, however, entails multiple calls of $\Pi_{LT}$ and $\Pi_{Mul}$, rendering it inefficient.


To solve the above problems, we design an efficient SMPC protocol $\Pi_{GeLU}$ based on segmented polynomials and \emph{Fourier series}. 
\begin{figure}[t]
	\centering\includegraphics[width=0.485\textwidth]{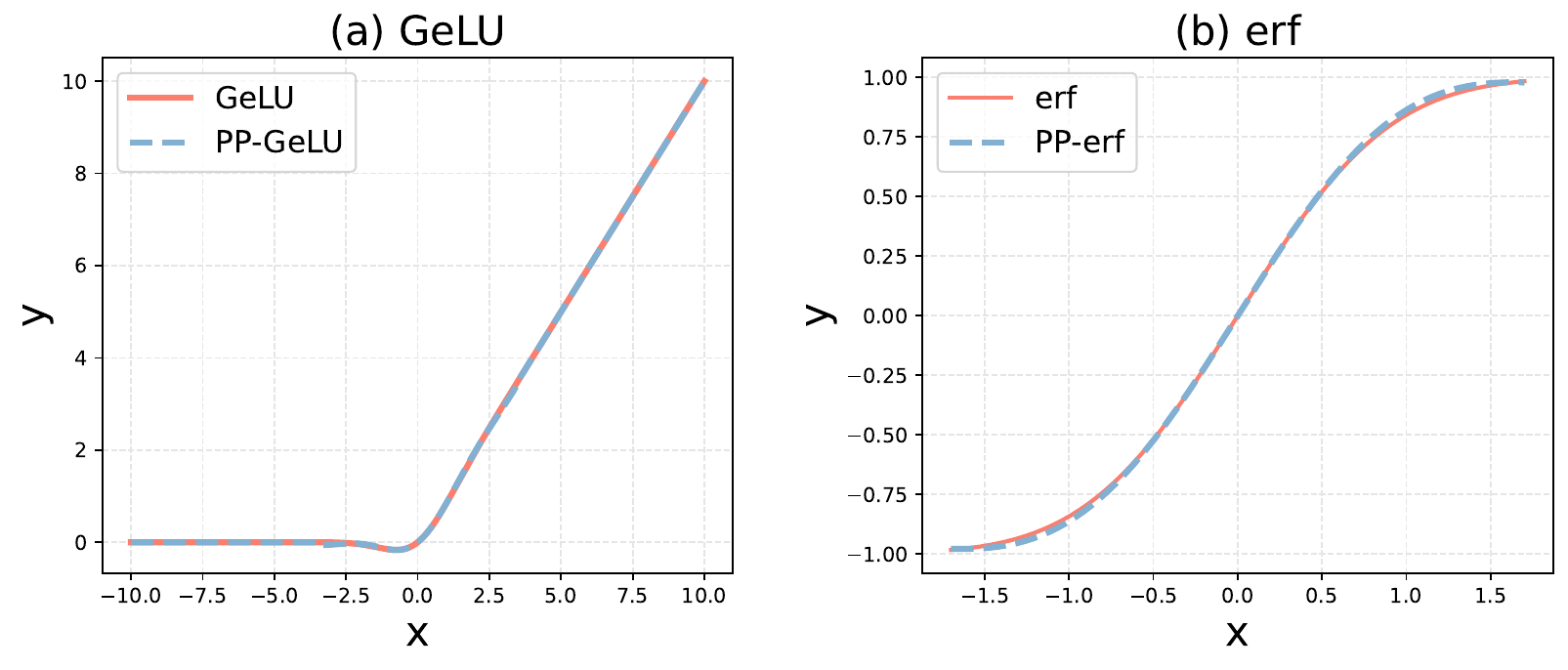}
	\caption{Fitting results of GeLU and erf functions.}
	\label{fig: GeLU & erf}
\end{figure}
As shown in~\cref{fig: GeLU & erf}, the erf function is an odd function symmetric about the origin with $\lim_{x\rightarrow \infty}\text{erf}(x) = 1$ and $\lim_{x\rightarrow -\infty}\text{erf}(x) = -1$. Therefore, we can convert it to the following segmented function
\begin{equation}\label{eq:erf}
    \text{erf}(x)\approx \left\{
	\begin{aligned}
	& -1,\quad & x < -1.7\\
	&f(x),\quad &-1.7 \leq x \leq 1.7\\
	&\quad1, \quad &x > 1.7\\
	\end{aligned}, 
	\right.
\end{equation}
where $f(x)$ can be approximated through a Fourier series composed of sine functions with a period\footnote{The results of the sine function fitting for different periods are shown in \cref{sec: fourier fitting}.} of $20$. Although a greater number of terms enhances the accuracy of the fitting outcomes, it concurrently leads to increased communication overhead. Here, we employ the following 7-term Fourier series:
\begin{equation}\label{eq: fur}
    f(x) = \bm{\beta} \odot \sin(\bm{k}\odot \pi x/10), 
\end{equation}
where $\bm{k} = (1,2,3,4,5,6,7)$, $\bm{\beta}$ is the  Fourier series coefficients and $\odot$ denotes the element-wise multiplication. For $i = 1,2,\dots,7$,  
\begin{equation}\label{eq: fur coefficients}
\bm{\beta}_i = \frac{1}{10}\int_{-10}^{10}\text{erf}(x)\sin(\frac{\bm{k}_i\pi x}{10})dx \ .
\end{equation}
According to \cref{eq: fur coefficients}, we can compute the coefficients $\bm{\beta}$ = $(1.25772$, $-0.0299154$, $0.382155$, $-0.0519123,$ $0.196033, -0.0624557, 0.118029)$. 

Based on~\eqref{eq:erf}, the computation of the erf function is converted into comparison and sine function. The precise calculation of GeLU can be accomplished by combining $\Pi_{Mul}$ with the erf function. The specific steps of the SMPC protocol for GeLU are shown in~\Cref{alg:pp-gelu}.
Specifically, in steps 1-5 of~\Cref{alg:pp-gelu}, we determine in which interval of the segmented function the input $x$ falls by calling the $\Pi_{LT}$. Then, in step 7, the privacy-preserving computation of $f(x)$ is achieved by utilizing the $\Pi_{sin}$ presented in \citep{zheng2023secure}.
The algorithm requires only $1$ round of communication, and the specific steps of it is in \cref{sec: nonlinear protocol}. In steps~8-10, we compute the erf function and execute the GeLU calculation by invoking $\Pi_{Mul}$.

\paragraph{Protocol for LayerNorm.}\label{sec:pp-norm}

Previous work~\citep{knott2021crypten} implements the privacy-preserving computation of LayerNorm in~\eqref{eq:norm} by sequentially invoking $\Pi_{rSqrt}$ and $\Pi_{Div}$, resulting in expensive computational and communication overheads. Goldschmidt's method enables the direct conversion of \emph{square root inverses} (i.e., $\frac{1}{\sqrt{.}}$) directly into multiple iterations of multiplications. However, achieving a broader convergence range often requires complex nonlinear initial value calculations, such as Look-up-table (LUT) \citep{rathee2021sirnn} or exponentiation \citep{knott2021crypten}, before the iteration begins. To resolve this issue, 
we propose to employ the \emph{deflation technique} for bypassing 
these intricate nonlinear initial value calculations that are incompatible with SMPC. The detailed steps of the SMPC protocol for LayerNorm are in Algorithm \ref{alg:pp-layernorm}.


Specifically, in steps 3-8, we use Goldschmid's method to compute $\frac{1}{\sqrt{q}}$ where $q = (var(\bm{x}) + \epsilon)/\eta$. Through division by a constant $\eta$ (A hyperparameter whose value is shown in \cref{sec: models and hyperparameter}.), we initially deflate the denominator into the interval $[0.001, 2.99]$ which ensures fast convergence for linear initial values. Then, we set the initial values $p_0 = 1$, $q_0 = q$, and compute $m_i = (3 - q_{i-1})/2, p_{i} = p_{i-1} m_{i}, q_{i} = q_{i-1} m_{i}^2$ at each iteration by calling $\Pi_{Mul}$ and $\Pi_{Square}$. After $t = 11$ iteration, the value of $\frac{1}{\sqrt{q}}$ is calculated. 



\begin{algorithm}\footnotesize
    \caption{SMPC Protocol for LayerNorm $\Pi_{LayerNorm}$}
    \label{alg:pp-layernorm}
    \LinesNumbered
    \SetNoFillComment
    \DontPrintSemicolon
    \KwIn{For $j \in \{0, 1\}$, $S_j$ holds the shares $[\bm{x}]_j$.}
    \KwOut{For $j \in \{0, 1\}$, $S_j$ holds the shares $[\bm{y}]_j$ with $\bm{y} = \text{LayerNorm}(\bm{x})$.}
    
    \tcc{Compute the mean and variance}
    $[\![\bar{x}]\!] = \frac{1}{n}\cdot \sum^{n}_{h=1}[\![x_h]\!]$\; $[\![var(\bm{x})]\!] =  \sum^{n}_{h=1}\Pi_{Square}([\![x_h]\!] - [\![\bar{x}]\!])$\;
    \tcc{Goldschmidt's method}
    \tikzmark{fedavg:begin} $p_0 = 1, q_0 = \frac{1}{\eta}([\![var(\bm{x})]\!] + \epsilon)$\; 
    \For{$i\leftarrow 1$ \KwTo $t$}{
        $[\![m_i]\!] \leftarrow  (3 - q_{i-1})/2$\;
        $[\![q_i]\!] \leftarrow  \Pi_{Mul}([\![q_{i-1}]\!], \Pi_{Square}([\![m_{i}]\!]))$\;
        $[\![p_i]\!] \leftarrow  \Pi_{Mul}([\![p_{i-1}]\!], [\![m_{i}]\!])$ 
        \qquad \qquad \tikzmark{fedavg:end}
    }
  \drawCodeBox{fill=my_red!80}{fedavg:begin}{fedavg:end}
  \tcc{Compute $\text{LayerNorm}(x)$}
  $[\![\bm{y}]\!] = \gamma \cdot (\frac{1}{\eta}([\![\bm{x}]\!]-[\![\bar{x}]\!])\cdot [\![p_t]\!]) + \beta$\;
\end{algorithm}

\paragraph{Protocol for Approximated Softmax.}\label{sec:pp-Softmax}




As mentioned in Section~\ref{sec: Overview}, we follow \citet{li2022mpcformer} and bypass 
the privacy-preserving computations of exponential and maximum by substituting Softmax with 2Quad in \eqref{eq: 2quad}. However, to preserve the normalized nature of Softmax, the division operations cannot be avoided and thus become a new obstacle. 

To solve this problem, we again use the Goldschmidt's method to convert the division operation to multiplications. Similar to the LayerNorm protocol, the complex calculation of initial values is avoided by effective deflation. The implementation of the SMPC protocol for the approximated Softmax (i.e.,  $\Pi_{2Quad}$) is shown in~\Cref{sec:pp-softmax}.


\section{Experiments}\label{sec: experiments}
This section showcases the effectiveness of SecFormer through extensive experiments. 
We begin with the experiment setup in \cref{sec: settings} and then report the performance assessment results in \cref{sec: performance} and efficiency evaluations in \cref{sec: efficiency}, respectively. 
We further provide more analysis for SecFormer in \cref{sec:ablationstudy}, including an efficiency evaluation for $\Pi_{GeLU}$, $\Pi_{LayerNorm}$ and $\Pi_{2Quad}$.



\subsection{Experimental Setup}\label{sec: settings}
\paragraph{Implementation.} 
We implemented SecFormer using CrypTen\footnote{https://github.com/facebookresearch/CrypTen}, a semi-honest privacy-preserving machine learning framework based on secret sharing. To simulate the experimental environment, we utilized three Tesla V100 servers with a 10GB/s bandwidth. The hyperparameters for model fine-tuning and distillation follow the settings in \citep{li2022mpcformer}, see \cref{sec: models and hyperparameter} for details.

\paragraph{Baselines.} 
We compare SecFormer with state-of-the-art works based on model design (MPCFormer \citep{li2022mpcformer}) and SMPC protocol design (PUMA \citep{dong2023puma}). Specifically, MPCFormer improves the efficiency of PPI by substituting Softmax and GeLU with some SMPC friendly quadratics. PUMA enhances PPI efficiency by designing more efficient SMPC protocols for GeLU, LayerNorm and Softmax. Following the setting in MPCFormer, we include the result of the fine-tuned redesigned model as the baseline, denoted as MPCFormer$_{w/o}$ and SecFormer$_{w/o}$ (i.e., fine-tuned without distillation). We also re-implement PUMA on CrypTen for consistency.  


\paragraph{Models and Datasets.} We followed MPCFormer using a representative transformer model BERT, see \cref{sec: models and hyperparameter} for details.
For the reliability of the experimental results, we use datasets with different evaluation metrics and sizes, including RTE, MRPC, CoLA, STS-B, and QNLI. In terms of evaluation metrics, MRPC uses F1 scores, STS-B employs the average of Person and Spearman correlations, CoLA uses Matthews correlations, and RTE and QNLI rely on accuracy.


\subsection{Performance Comparison} \label{sec: performance}


We validate the performance of SecFormer and the main results are shown in \cref{tab: performance result}. For the model design framework MPCFormer, SecFormer exhibits a significant performance improvement. Specifically, for BERT$_{\text{BASE}}$, SecFormer outperforms MPCFormer across all tasks, resulting in a $3.4\%$ average improvement. For BERT$_{\text{LARGE}}$, MPCFormer faces significant performance degradation, including CoLA task failure. In contrast, even without data distillation, SecFormer outperforms MPCFormer. After distillation, SecFormer demonstrates a substantial $24.7\%$ performance improvement compared to MPCFormer. This is mainly because SecFormer implements an accurate computation of GeLU instead of replacing it aggressively with a quadratic polynomial. 

For the protocol design framework PUMA, SecFormer incurs only a marginal $0.9\%$ and $1.3\%$ performance degradation. PUMA does not perform any model design and achieves PPI without performance loss. However, this results in PUMA facing unacceptable communication overheads, as detailed in \cref{sec: efficiency}.


     
    
    

\begin{table*}[ht]
    \centering
    \resizebox{0.8\linewidth}{!}{
    \begin{tabular}{c|ccc|c|c|c|c|c|c}
    \toprule
     \multirow{2}{*}{Models} & \multirow{2}{*}{Methods} & GeLU & Softmax & QNLI& CoLA &STS-B & MRPC & RTE & \multirow{2}{*}{\textbf{Avg.}} \\
    & & Approx.& Approx.& (108k)  &  (8.5k)  & (5.7k) &  (3.5k) &  (2.5k) &  \\
    \midrule  
     \multirow{6}{*}{BERT$_{\text{BASE}}$}  &  Plain-text &  GeLU & Softmax & $91.7$   & $57.8$ & $89.1$ &  $90.3$ & $69.7$ & \boldmath{$79.7$}\\
    \cmidrule(lr){2-10}
      & PUMA &  GeLU & Softmax & $91.7$  & $57.8$ & $89.1$ & $90.3$ & $69.7$ & $79.7^*$\\
    \cmidrule(lr){2-10}
    & MPCFormer$_{w/o}$ & Quad & 2Quad & $69.8$   & $0.0$ & $36.1$ & $81.2$ & $52.7$ &  $48.0$\\
   & MPCFormer  & Quad & 2Quad & $90.6$   & $52.6$ & $80.3$ & $88.7$ & $64.9$&  $75.4$ \\
      \cmidrule(lr){2-10}
    & SecFormer$_{w/o}$ &  GeLU & 2Quad  & $89.3$  & $57.0$  & $86.2$ & $83.8$ & $63.2$ & $75.9$\\
    
     & SecFormer & GeLU & 2Quad & $91.2$   & $57.1$ & $87.4$ & $89.2$  &$69.0$& $78.8^*$\\
     \midrule
      \multirow{6}{*}{BERT$_{\text{LARGE}}$} & Plain-text &  GeLU & Softmax &  $92.4$  & $61.7$ & $90.2$ & $90.6$ & $75.5$ & \boldmath{$82.1$}\\
    \cmidrule(lr){2-10} 
       & PUMA &  GeLU & Softmax & $92.4$ & $61.7$  & $90.2$ & $90.6$&  $75.5$ & $82.1^*$ \\
    \cmidrule(lr){2-10}
    & MPCFormer$_{w/o}$ & Quad & 2Quad & $49.5$ & $0.0$ & $0.0$ & $81.2$& $52.7$ & $36.7$\\
    & MPCFormer  & Quad & 2Quad & $87.8$  & $0.0$ & $52.1$ &$81.4$ &$59.2$& $56.1$  \\
    
       \cmidrule(lr){2-10}
    & SecFormer$_{w/o}$ &  GeLU & 2Quad  & $90.8$ &  $60.8$ &  $89.0$ & $87.6$ &$69.7$ & $79.6$\\
    
     & SecFormer & GeLU & 2Quad &$92.0$  &$61.3$  & $89.2$ & $88.7$ & $72.6$ & $80.8^*$\\
    \bottomrule
    \end{tabular}}
    \caption{Performance comparison of BERT$_{\text{BASE}}$ and BERT$_{\text{LARGE}}$. Bolded numbers indicate best results; numbers marked ``*'' indicate performance loss within $1.5\%$. For BERT$_{\text{BASE}}$, we directly use the results from \citep{li2022mpcformer}. For BERT$_{\text{LARGE}}$, \citet{li2022mpcformer} uses the 2ReLU instead of 2Quad for performance reasons, where $\text{2ReLU}(\bm{x})[i] = \text{ReLU}(\bm{x})[i]/\sum^{n}_{h=1}\text{ReLU}(\bm{x})$. Calculating ReLU requires a call to $\Pi_{LT}$. This results in more overhead than calculating 2Quad.} 
    \label{tab: performance result}
\end{table*}

\subsection{Efficiency Comparison} \label{sec: efficiency}
We evaluate the efficiency by testing the time and communication volume required to perform single-sample inference across different frameworks. The main results are shown in \cref{tab:efficiency}. 
We can find that SecFormer is significantly more efficient than PUMA. Specifically, for BERT$_{\text{BASE}}$ and BERT$_{\text{LARGE}}$, SecFormer performs $3.57$ and $3.58$ faster than PUMA on the total inference time. These advantages stem from that SecFormer utilizes model design to achieve efficient computation of Softmax, and design efficient SMPC protocols suitable for the Transformer models for other nonlinear operators (i.e., GeLU, LayerNorm) by using appropriate numerical computation techniques. The efficiency of each SMPC protocol is shown in \cref{tab:efficiency} and will be discussed later in \cref{sec:ablationstudy}.

When considering the framework of model design, SecFormer is only $1.05$ and $1.04$ times slower than MPCFormer in the scenarios of BERT$_{\text{BASE}}$ and BERT$_{\text{LARGE}}$, respectively. This result is based on the fact that SecFormer spends 41\% of its time performing privacy-preserving calculations for GeLU, while MPCFormer spends only 0.01\% of its time to implement the privacy-preserving calculations for Quad. However, the conclusions in \cref{sec: performance} suggest that replacing GeLU with quadratic leads to dramatic degradation of model performance or even failure on some tasks (i.e., performance with $0$ in Table~\ref{tab: performance result}).


In conclusion, experiments with SecFormer regarding performance and efficiency reveal its dual advantages, combining strengths from both protocol design and model design frameworks.

    
    

\begin{table*}[htbp]
    \centering
    \resizebox{\linewidth}{!}{
    \begin{tabular}{c|cc|c|c|c|c|c|c|c|c}
    \toprule
     \multirow{2}{*}{Models} & \multirow{2}{*}{Methods}  & \multicolumn{2}{c}{GeLU} &\multicolumn{2}{c}{Softmax} &\multicolumn{2}{c}{LayerNorm} & \multicolumn{2}{c}{Others} & Total\\
     & & Times(s)& Comm(GB)& Times(s)& Comm(GB) & Times(s)& Comm(GB) & Times(s)& Comm(GB)& Times(s)\\
    \midrule
    \multirow{6}{*}{BERT$_{\text{BASE}}$} & CrypTen &  $16.46$&  $28.689$ & $37.25$ & $50.285$ & $6.614$&  $0.492$&  $9.365$  & $3.463$ & $71.097$\\
    \cmidrule{2-11}  
       & PUMA  & $16.343$& $28.689$ & $42.219$ & $67.837$ & $2.285$&  $0.477$& $8.781$   & $3.463$ & $69.661$ \\
    \cmidrule{2-11}
   & MPCFormer   &  $0.351$& $0.604$ & $3.129$ & $1.895$ & $6.522$&  $0.497$ & $8.589$   & $3.463$ & \boldmath{$18.591$}  \\
       \cmidrule{2-11}
     & SecFormer & $8.132$& $17.817$ & $1.362$ & $1.844$ & $1.523$&  $0.468$& $8.496$   & $3.463$ & $19.513^*$\\
         \midrule
      \multirow{6}{*}{BERT$_{\text{LARGE}}$} & CrypTen & $27.881$& $57.378$ & $83.017$ & $134.093$ & $9.105$&  $1.272$& $19.945$  & $8.565$ & $140.018$\\
    \cmidrule{2-11}  
       & PUMA  &  $27.357$& $57.378$ & $89.938$ & $180.898$ & $4.313$& $1.254$ & $18.278$   & $8.565$ &$139.954$ \\
    \cmidrule{2-11}
    & MPCFormer  &  $0.351$& $0.604$ & $7.274$ & $5.052$ & $10.864$&  $1.282$& $19.261$   & $8.565$ & \boldmath{$37.75$}  \\
       \cmidrule{2-11}
     & SecFormer &  $14.531$ & $35.635$ & $3.115$ & $4.916$ & $3.122$&  $1.248$ & $18.321$  & $8.565$ & $39.089^*$\\
     \bottomrule
    \end{tabular}}
    \caption{Efficiency Comparison of BERT$_{\text{BASE}}$ and BERT$_{\text{LARGE}}$. Bolded numbers indicate the best results; Numbers marked ``*'' indicate within 2 seconds slower than the best result. The results are the average of ten runs.}
    \label{tab:efficiency}
\end{table*}

\begin{figure}[t]
	\centering
        \includegraphics[width=0.485\textwidth]{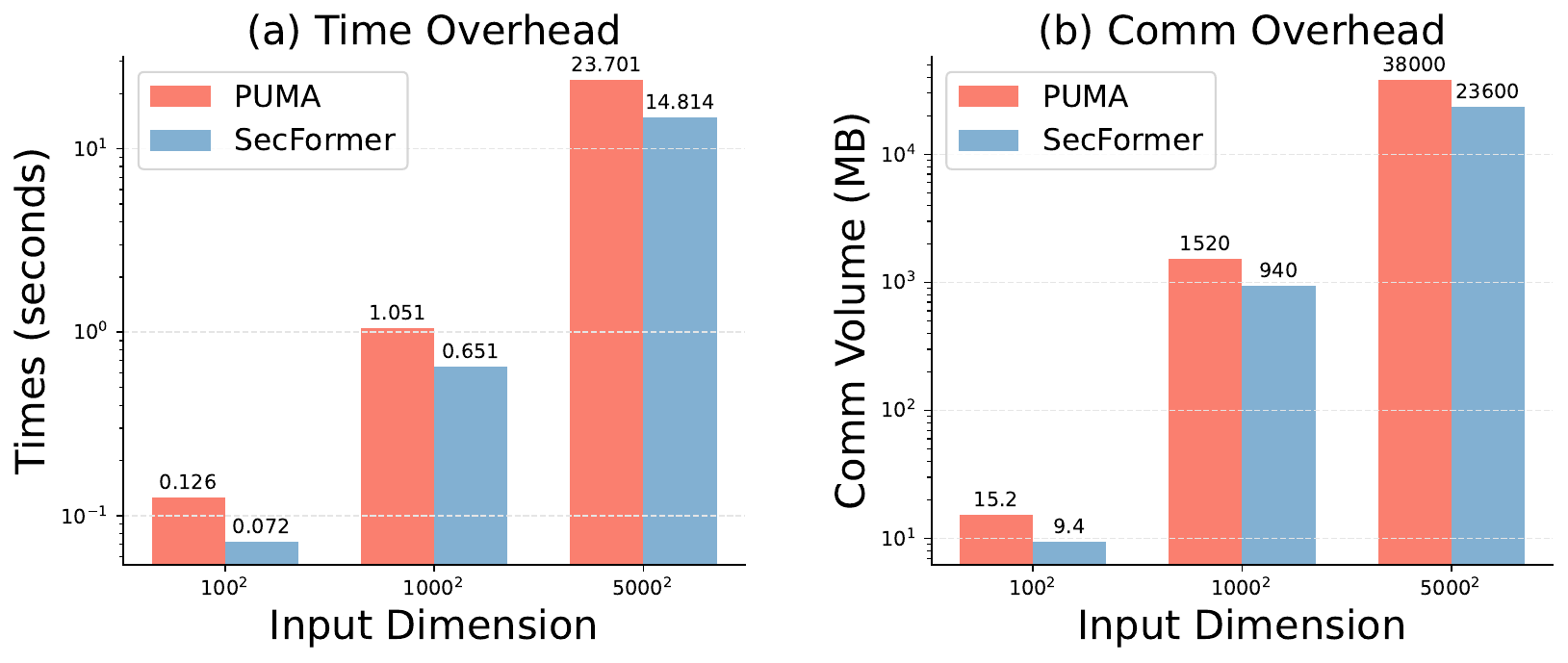}
	\caption{Comparison of $\Pi_{GeLU}$ Time and Communication Overheads.}
	\label{fig: PP-glue}
\end{figure}

\subsection{SMPC Protocols Evaluation}\label{sec:ablationstudy}
We compare $\Pi_{GeLU}$ with PUMA in terms of time and communication overhead. The comparison results in \cref{fig: PP-glue} show that $\Pi_{GeLU}$ is about 1.6 times lower than PUMA in time and communication overhead. This is mainly due to the fact that it invokes fewer $\Pi_{LT}$ relative to PUMA. In terms of accuracy, both $\Pi_{GeLU}$ and PUMA meet the needs of PPI, while CrypTen can only maintain accuracy over a small range. See \cref{sec: Accuracy of ppgelu} for details.

\begin{figure}[t]
	\centering
	\includegraphics[width=0.485\textwidth]{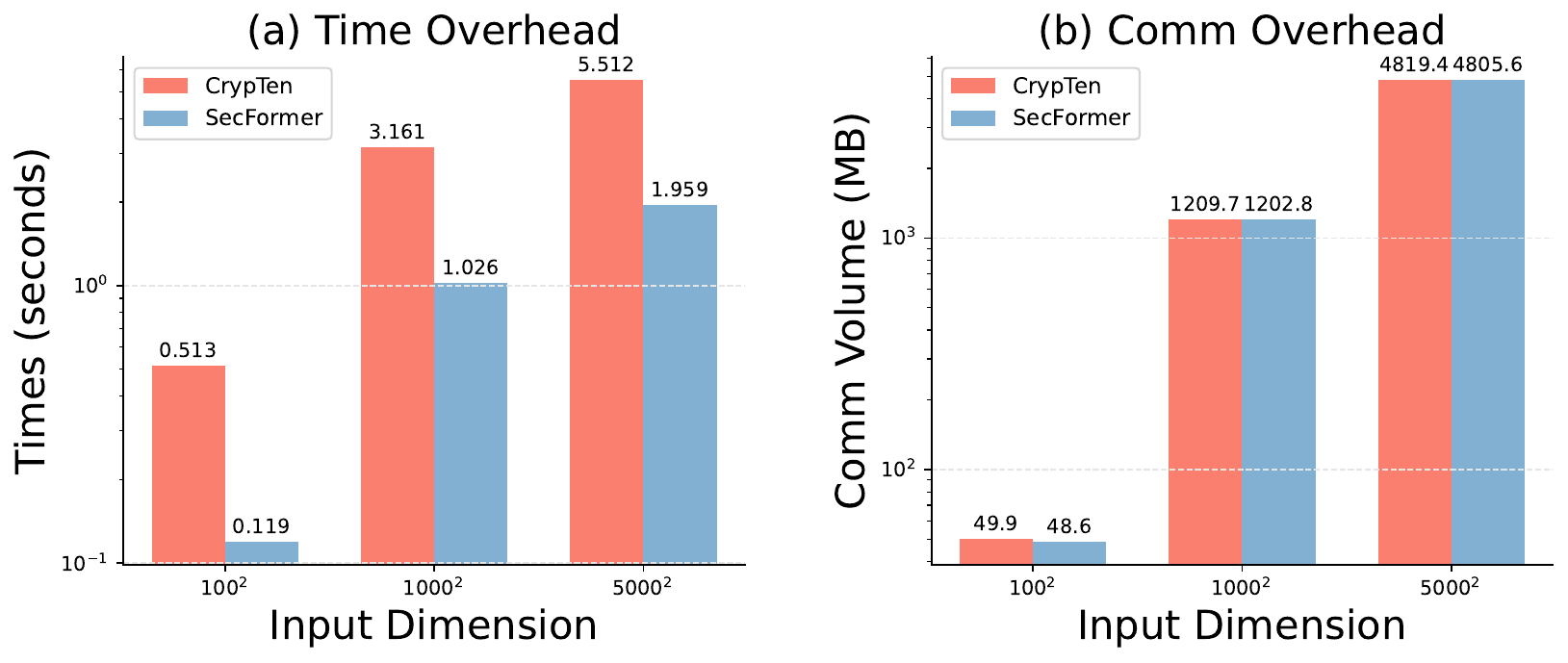}
	\caption{Comparison of $\Pi_{LayerNorm}$ Time and Communication Overheads.}
	\label{fig: pp-norm}
\end{figure}
\begin{figure}[t]
	\centering
       \includegraphics[width=0.485\textwidth]{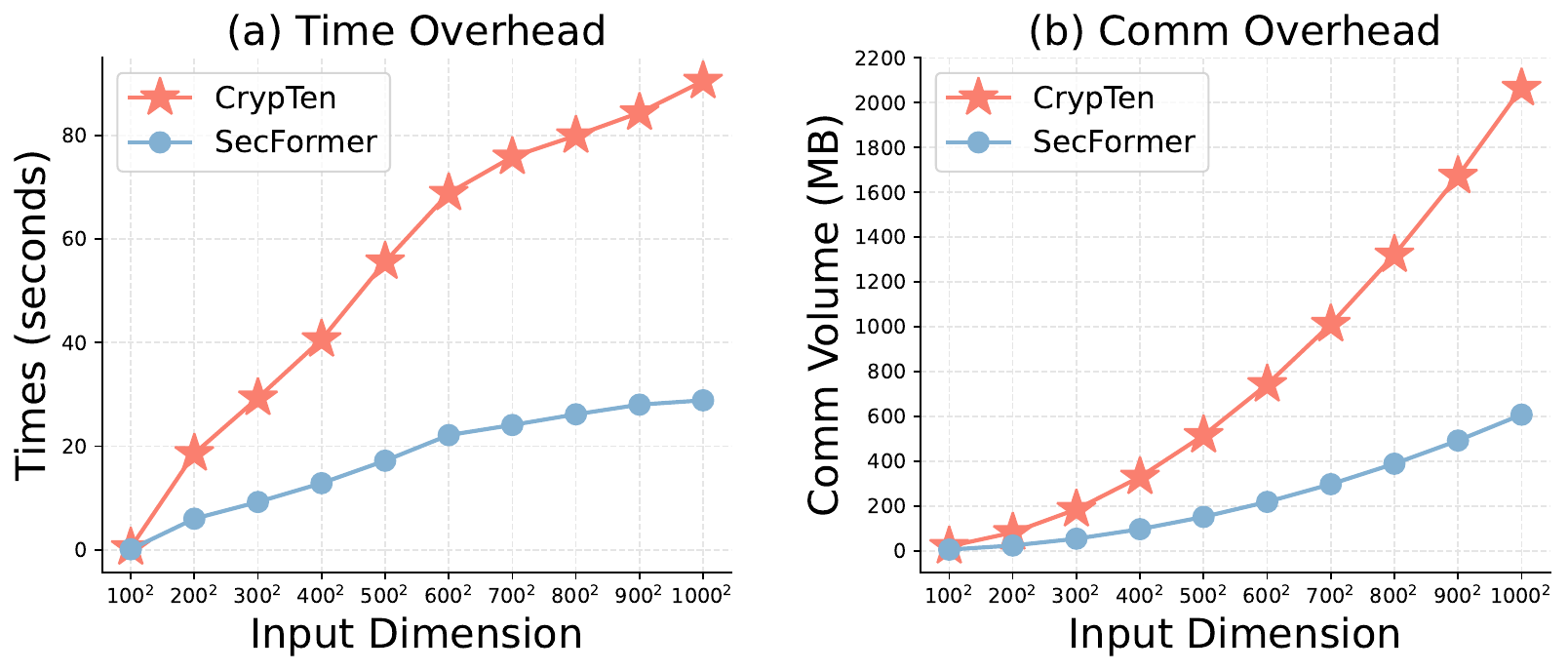}
	\caption{Comparison of Privacy-preserving Calculation for Square-root Inverse Time and Communication Overheads.}
	\label{fig: inv_sqrt}
\end{figure}

We compare $\Pi_{LayerNorm}$ with CrypTen~\citep{knott2021crypten} in terms of time and communication overhead. \cref{fig: pp-norm} shows that $\Pi_{LayerNorm}$ is up to 4.5 times faster than CrypTen \citep{knott2021crypten}. This is due to the efficient privacy-preserving square root inverse calculation proposed by SecFormer. As shown in \cref{fig: inv_sqrt}, it is 4.2 times faster than CrypTen and reduces the communication volume by a factor of 2.5. 

\begin{figure}[ht]
	\centering
	\includegraphics[width=0.485\textwidth]{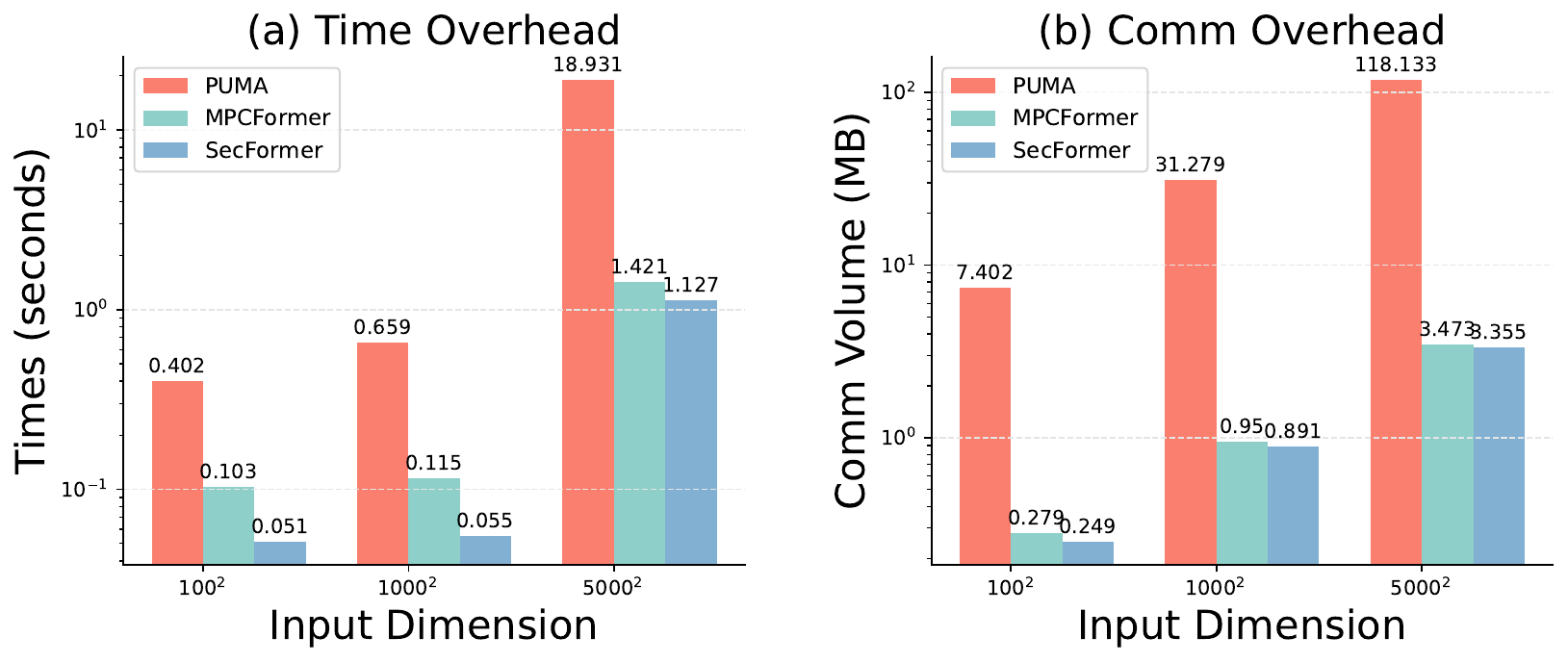}
	\caption{Comparison of $\Pi_{2Quad}$ Time and Communication Overheads.}
	\label{fig: pp-softmax}
\end{figure}

\begin{figure}[ht]
	\centering
       \includegraphics[width=0.485\textwidth]{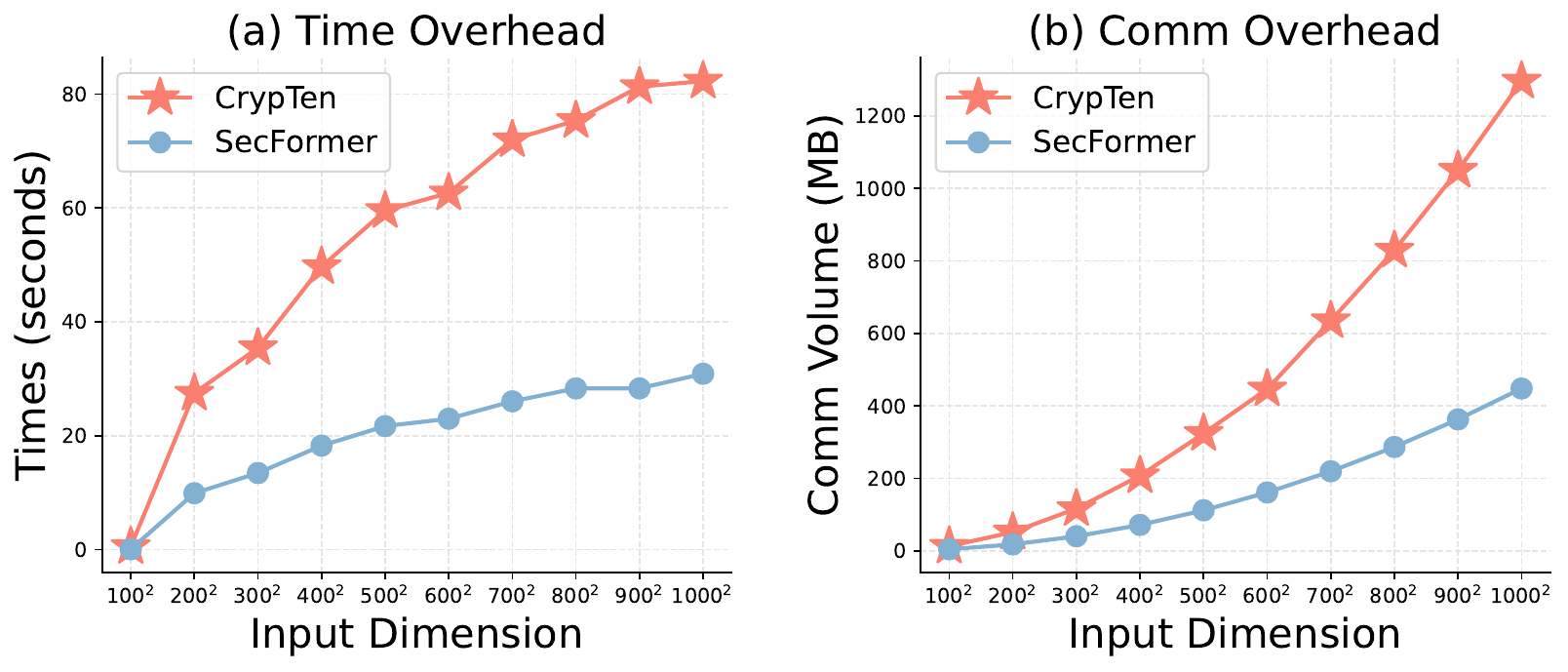}
	\caption{Comparison of Privacy-Preserving Division Calculation Time and Communication Overheads.}
	\label{fig: pp-div}
\end{figure}

We compare $\Pi_{2Quad}$ with MPCFormer and PUMA in terms of time and communication overhead. \cref{fig: pp-softmax} shows that $\Pi_{2Quad}$ is $1.26 \sim 2.09$ times faster than MPCFormer and the communication overhead is reduced by $1.04 \sim 1.12$ times. These enhancements come from the efficient privacy-preserving division calculation proposed by SecFormer.
As shown in \cref{fig: pp-div}, it is 3.2 times faster than CrypTen, and the communication overhead is reduced by 1.6 times.

Compared to PUMA, which achieves precise privacy-preserving Softmax, $\Pi_{2Quad}$ gets a drastic improvement in efficiency, i.e., $8.24 \sim 16.8$ times faster and $30.53 \sim 36.2$ times less communication. This is due to the fact that the model design performed by SecFormer avoids the calculation of exponential and maximum.
\section{Conclusion}

We present SecFormer, a synergistic PPI framework that strategically combines the strengths of both SMPC protocol design and Transformer model design. Extensive experiments reveal that SecFormer surpasses existing PPI methods, achieving fast and accurate PPI for Transformer models. It not only matches the performance of approaches that focus exclusively on SMPC protocols but also competes with the efficiency of methods dedicated solely to model design. 
SecFormer holds significant potential for enhancing large language models, offering 
an effective solution that promises to maintain high performance while ensuring stringent privacy and efficiency standards in increasingly complex and expansive linguistic landscapes.

\section{Limitations}
We summarize the limitations of SecFormer as follows: 
(1) SecFormer focuses on implementing PPI for the encoder-only Transformer model, such as BERT, without extending to other Transformer model families like the GPT series. We concentrate on the encoder-only Transformer model because of its continued prominence in real-world natural language understanding tasks, particularly within resource-constrained environments like edge computing. Prior efforts to implement the encoder-only Transformer model for PPI have encountered obstacles, including slow inference speeds and substantial performance degradation. Our work addresses these challenges and offers insights to guide future optimization efforts concerning PPI across diverse Transformer model families. The proposed protocols can be applied to implement PPI of other transformer-based models straightforwardly and we will consider PPI for decoder only Transformer models like GPT in the future.
(2) Regarding SMPC protocols, SecFormer executes only on CrypTen and does not invoke the cutting-edge underlying SMPC protocols. We will try to exploit other privacy-preserving frameworks with more advanced SMPC protocols to further improve the inference efficiency of SecFomer in future work. 
(3) SecFormer only performs model design by replacing Softmax with 2Quad and does not incorporate other model lighting techniques.
Other model lightweight techniques such as model quantization and pruning are compatible with the proposed SMPC protocols and can be combined into SecFormer to further improve the PPI efficiency in the future.
\section{Acknowledgements}\label{sec:Acknowledgements}
This research is partially supported by the National Key R\&D Program of China (No.2021ZD0112905), the National Natural Science Foundation of China (No. 62206139, 62106114), and the Major Key Project of PCL (PCL2023A09).
\bibliography{ref}
\clearpage
\appendix

\section{2-out-of-2 Secret Sharing}\label{sec: secret sharing}
The 2-out-of-2 secret sharing includes arithmetic secret sharing and Boolean secret sharing. The 2-out-of-2 arithmetic secret sharing contains two algorithms: 
\begin{itemize}[leftmargin=*, itemsep=0pt, labelsep=5pt]
    \item $Shr(x) \rightarrow ([x]_0, [x]_1)$ is used to generate the shares by randomly selecting a number $r$ from $\mathcal{Z}_L$, letting $[x]_0=r$, and computing $[x]_1=(x - r)\mod L$;
    \item $Rec([x]_0, [x]_1)\rightarrow x$ is used to reconstruct the original value from the shares, which can be done by simply calculating $([x]_0 + [x]_1) \mod L$. 
\end{itemize}

Note that due to the randomness of $r$, neither a single $[x]_0$ nor $[x]_1$ can be used to infer the original value of $x$. The arithmetic secret sharing technique has been widely used to construct SMPC protocols for ML operations (e.g., $+$, $-$ and $\cdot$, etc.) such that both the inputs and outputs of the protocol are the arithmetic shares of the original inputs and outputs:
\begin{equation}
    \Pi_f([inputs]_0, [inputs]_1) \rightarrow ([f]_0, [f]_1), 
\end{equation}
where $\Pi_f$ denotes an SMPC protocol of the operation $f$. The shares in $\mathcal{Z}_2$ is called \emph{Boolean} shares, and the operations of $+$, $-$ and $\cdot$ are replaced by bit-wise operations $\oplus$ and $\wedge$. We use $[\![x]\!]$, $\langle\!\langle x \rangle\!\rangle$ denotes the arithmetic and boolean shares of $x$, i.e., $[\! [x]\! ] = ([x]_0, [x]_1), \langle\!\langle x \rangle\!\rangle = (\langle x\rangle_0, \langle x\rangle_1)$.

\section{Protocol for the Approximated Privacy-Preserving Softmax}\label{sec:pp-softmax}
In this section, we give the specific implementation of the SMPC Protocol for the approximated Softmax (i.e., 2Quad) as mentioned in \cref{sec: Overview}. In steps 3-8 of~\Cref{alg:pp-Softmax}, we first deflate the denominator $q = \sum^{n}_{h=1}(\bm{x} + c)^2$ into the interval $ [0.001,1.999]$, which ensures fast convergence for linear initial values, through division by a constant $\eta$. 
Subsequently, we set the initial values $q_0 = q, p_0 = (\bm{x} + c)^2$, and compute $m_i = 2 - q_{i-1}, p_{i} = p_{i-1} m_i, q_{i} = q_{i-1} m_i$ at each iteration by calling $\Pi_{Mul}$. After $t=13$ iterations, $\frac{p}{q}$ is computed. 


\begin{algorithm}\footnotesize
    \caption{SMPC Protocol for Softmax $\Pi_{2Quad}$}
    \label{alg:pp-Softmax}
    \LinesNumbered
    \SetNoFillComment
    \DontPrintSemicolon
    \KwIn{For $j \in \{0, 1\}$, $S_j$ holds the shares $[\bm{x}]_j$.}
    \KwOut{For $j \in \{0, 1\}$, $S_j$ holds the shares $[\bm{y}]_j$, where $\bm{y} = 2quad(\bm{x})$.}
    \tcc{Compute the numerator}
$[\![\bm{p}]\!] = \Pi_{Square}([\![\bm{x} + c]\!])$\;
    \tcc{Compute the denominator}
   $[\![q]\!] = \sum^{n}_{h=1}[\![\bm{p}[h]]\!]$\;
    \tcc{Goldschmidt's method}
$q_0 = \frac{1}{\eta}[\![q]\!], [\bm{p}_{0}]\!] = \frac{1}{\eta}[\![\bm{p}]\!]$\; 
 \For{$i\leftarrow 1$ \KwTo $t$}{
  $[\![m_{i}]\!] \leftarrow  2 - [\![q_{i-1}]\!]$\;
   $[\![\bm{p_i}]\!] \leftarrow  \Pi_{Mul}([\![\bm{p}_{i-1}]\!], [\![\bm{m}_{i}]\!])$\;
     $[\![q_{i}]\!] \leftarrow  \Pi_{Mul}([\![q_{i-1}]\!], [\![m_{i}]\!])$\;}
$[\![\bm{y}]\!] = [\![\bm{p_t}]\!]$\;
\end{algorithm}

\section{Accuracy Comparison of Privacy-Preserving GeLU Algorithms}\label{sec: Accuracy of ppgelu}
In this section we compare the performance of privacy-preserving GeLU with Puma and CrypTen. The specific comparison results are shown in \cref{tab:acc_gelu}. Both SecFormer and Puma achieve privacy-preserving computation within the entire interval of the GeLU function by using segmented polynomials. CrypTen, on the other hand, locally fits the erf function using a low-order Taylor expansion and thus can only achieve privacy-preserving computation of the GeLU function in a smaller interval.

\begin{table*}[htbp]
    \centering
    \resizebox{\linewidth}{!}{
    \begin{tabular}{c|ccc|ccc|ccc}
    \toprule
     \textbf{ Input Interval }& \multicolumn{3}{c|}{$[-1,1]$} & \multicolumn{3}{c|}{$[-5,5]$} & \multicolumn{3}{c}{$[-10,10]$}\\
    \midrule
       \textbf{Methods}  & CrypTen & Puma & SecFormer & CrypTen & Puma & SecFormer & CrypTen & Puma & SecFormer\\
    \midrule  
       \textbf{Error Mean } & 0.001 & 0.005 & 0.001 & 30437.717 & 0.003 & 0.005 & 7480209.5 & 0.002 & 0.003\\
    \midrule
       \textbf{Error Var}  &$\pm8.37\times10^{-6}$ & $ \pm 6.85\times10^{-6}$ & $\pm2.03\times10^{-6}$ & $\pm 3.28\times10^{9}$ &  $\pm 1.01\times10^{-5}$&  $\pm3.82\times10^{-5}$ &  $\pm 1.68\times10^{14}$ &  $\pm 7.06\times10^{-6}$ & $\pm 2.54\times10^{-5}$\\
    \bottomrule
    \end{tabular}}
    \caption{Accuracy Comparison of Privacy-Preserving GeLU Algorithms.}
    \label{tab:acc_gelu}
\end{table*}

\section{Security Proof and Communication Complexity Analysis}\label{sec: security proof & comm complexity}
\subsection{Security Proof}
SecFormer adheres to a semi-honest (also known as honest-but-curious) assumption similar to the works of \citet{li2022mpcformer} and \citet{ dong2023puma}, where honest participants constitute the majority. Under this assumption, the security of SecFormer can be formally proved within the simulation paradigm, specifically against static semi-honest adversaries denoted as $\mathcal{A}$, which can potentially corrupt one of the servers. The simulation paradigm delineates two distinct worlds: the real world and the ideal world. In the real world, the servers execute the protocols in the presence of semi-honest adversaries $\mathcal{A}$. In contrast, the ideal world involves the servers transmitting inputs to a trusted dealer capable of correctly executing the protocol. The security of SecFormer necessitates that, for any semi-honest adversary $\mathcal{A}$, the distribution of the real world remains indistinguishable from that of the ideal world. The definition of privacy-preserving inference protocols \citep{mishra2020delphi, HuangLHD22, hao2022iron} is as follows:

\begin{definition}\label{def: def1}
A protocol $\Pi_{P}$ between the servers who have the shares of the model weights and the inference data is a privacy-preserving protocol if it complies with the following criteria: (1) Correctness: For a model M with weights w and input samples x, the client's output at the end of the protocol is the correct inference M (w, x); and (2) Security: For a computational server $S_j, j \in \{0,1\}$ that is corrupted by adversary $\mathcal{A}$, there exists a probabilistic polynomial time simulator $Sim_{S_j}$ such that adversary $\mathcal{A}$ cannot distinguish $View^{\Pi_{P}}_{S_j}$ (i.e., the view of $S_j$ during the implementation of $\Pi_{P}$) from $Sim_{S_j}$. Similarly, for a corrupted server $T$, there exists an efficient simulator $Sim_T$ such that $View^{\Pi_{P}}_{T}$ is indistinguishable from $Sim_T$.
\end{definition}

SecFormer is constructed from the sub-protocols outlined in the works of \citet{knott2021crypten} and \citet{zheng2023secure}. Leveraging the concept of universally composable security established by \citet{canetti2001universally}, we can prove that SecFormer satisfies \cref{def: def1} directly.

\subsection{Communication Complexity Analysis}\label{sec: comm}
The execution of $\Pi_{GeLU}$ invokes two $\Pi_{LT}$, one $\Pi_{Sin}$ and one $\Pi_{Mul}$.
Thus the execution of the privacy-preserving GeLU algorithm takes a total of $2\log^{L} + 4$ rounds of online communication and transmit 7210 bits. 

The implementation of privacy-preserving LayerNorm requires calls to $\Pi_{Mul}$, $\Pi_{Square}$ and privacy-preserving inverse of the square root. The inverse of the square root requires one call to $\Pi_{Square}$ and two calls to $\Pi_{Mul}$ in parallel per iteration, costing 2 rounds of communication and transferring 640 bits. Thus performing the square root inverse takes a total of 22 rounds of communication and transfers 7040 bits and the implementation of privacy-preserving LayerNorm takes a total of 24 rounds of communication and transfers 7424 bits.

The implementation of approximate privacy-preserving Softmax requires calls to $\Pi_{Mul}$ and $\Pi_{Div}$. The $\Pi_{Div}$ requires two call to $\Pi_{Mul}$ in parallel per iteration, costing 1 rounds of communication and transferring 512 bits. Thus performing the $\Pi_{Div}$ takes a total of 13 rounds of communication and transfers 6,656 bits and the implementation of approximate privacy-preserving Softmax takes a total of 23 rounds of communication and transfers 6,784 bits.

\section{Underlying SMPC Protocols}\label{sec: underlying protocols}
In this section, we provide a brief overview of the underlying protocols used and refer to the works of \citet{knott2021crypten} and \citet{ zheng2023secure} for details.
Let ${S_j}$ with $j \in \{0,1\}$ be two parties that are used to execute the SMPC protocol. Each party $S_j$ will be given one additive share $([u]_j, [v]_j)\in \mathcal{Z}_L$ of the operation inputs $u$ and $v$ for $j \in \{0,1\}$.

\subsection{Privacy-Preserving Linear Protocols}\label{sec: linear protocol}

\paragraph{Privacy-preserving addition} is implemented with $[u + v]_j = [u ]_j + [v]_j$ for $j \in \{0, 1\}$.

\paragraph{Privacy-preserving multiplication} is implemented with Beaver-triples: $(a,b,c)$ where $a, b \in \mathcal{Z}_{L}$ are randomly sampled from $\mathcal{Z}_{L}$ and $c = a \cdot b \mod L$. Specifically, for each $j \in \{0, 1\}$,  $S_j$ first calculates $[d]_j = [u]_j-[a]_j$ and $[e]_j = [v]_j - [b]_j$. Then, they send the $[d]_j$ and $[e]_j$ to each other and reconstruct $d= Rec([d]_0, [d]_1)$ and $e =  Rec([e]_0, [e]_1)$. Finally, the additive share of $u \cdot v$ can be computed using $[u\cdot v]_j = -jd \cdot e +[u]_j \cdot e + d \cdot [v]_j + [c]_j$.
To complete the SS-based multiplication, both parties need to spend $1$ round of two-way communication and transmit 256 bits.

\subsection{Privacy-Preserving Non-Linear Protocols}\label{sec: nonlinear protocol}
\paragraph{Privacy-preserving comparison} is implemented by the conversion between the additive shares and the binary shares. Specially, $[\![z]\!] = [\![x - y]\!]$ is converted to the binary shares $\langle\!\langle z \rangle\!\rangle$ through additive circuit with $\log^{L}$ round of communication. Subsequently, the binary shares of $z$'s sign bit can be determined by $\langle\!\langle b \rangle\!\rangle =  \langle\!\langle z \rangle\!\rangle >> (l-1)$\footnote{$>>l$ denote shift $l$ bit to the right.}. Finally, the additive shares of $x < y$ can be derived by converting $\langle\!\langle b \rangle\!\rangle$ to $[\![b]\!]$ with one round of communication. Thus, the implementation of privacy-preserving compare algorithm cost $\log^{L} + 1$ round of communication and transmit 3456 bits.

\paragraph{Privacy-preserving maximum} of the n-element vector $\bm{x}$ is implemented by calling $\log^{n}$ privacy-preserving comparisons using the tree reduction algorithm \citep{knott2021crypten}.

\paragraph{Privacy-preserving exponential} is implemented using the repeated-squaring method 

\begin{equation}
    e^x = \text{lim}_{x \rightarrow \infty}\big (1 + \frac{x}{2^n}\big)^{2^n},
\end{equation}
which converts exponential calculations into addition and square operations. The number of iterations $n$ is set to 8 in \citep{knott2021crypten} by default.

\paragraph{Privacy-preserving reciprocal} is implemented by Newton-Raphson method, which converts reciprocal calculations into addition and multiplication operations. The iterative formula is
\begin{equation}
    y_{n+1} = y_n(2-xy_n).
\end{equation}
The initial value of the iteration is 
\begin{equation}
    y_{0} = 3e^{\frac{1}{2}-x}+0.003.
\end{equation}
The number of iterations is set to 10 in \citep{knott2021crypten} by default.

\paragraph{Privacy-preserving square root}
is implemented by Newton-Raphson method, which converts exponential calculations into addition and multiplication operations. The iterative formula is
\begin{equation}
    y_{n + 1} = \frac{1}{2}y_n(3-xy_n^2).
\end{equation}
The initial value of the iteration is 
\begin{equation}
    y_{0} = e^{-2.2(\frac{x}{2}+0.2)} + 0.198046875.
\end{equation}
The number of iterations is set to 3 in \citep{knott2021crypten} by default.

\paragraph{Privacy-preserving sine }
is implemented on trigonometric identities. Specifically, $\sin(x) = \sin(\delta)\cos(t) + \cos(\delta)\sin(t)$, where $\delta = x - t$. With the assistance of the server $T$, the random numbers $t, \sin(t), \cos(t)$ are generated in the offline phase, and the share of $\sin(x)$ is computed in the online phase with only one round of communication and transmits 42 bit.\footnote{CrypTen uses 16-bit computational precision.} See \Cref{alg:pp-sine} for an implementation of the privacy-preserving sine.

\begin{algorithm}\footnotesize
    \caption{Privacy-preserving sine}
    \label{alg:pp-sine}
    \LinesNumbered
    \SetNoFillComment
    \DontPrintSemicolon
    \KwIn{For $j \in \{0, 1\}$, $S_j$ holds the shares $[x]_j$; Same Pseudo-Random Function (PRF) and key $k_j$.}
    \KwOut{For $j \in \{0, 1\}$, $S_j$ returns the shares $[y]_j$, where $y = sin(x)$.}
    \tcc{Offline Phase}
    $S_0, T : [t]_0, [u]_0, [v]_0 \leftarrow PRF(k_0)$\;
    $S_1, T : [t]_1 \leftarrow PRF(k_1)$\;
    $T : t = [t]_0 + [t]_1,  [u]_1 = sin(t) -  [u]_0,  [v]_1 = cos(t) -  [v]_0$\;
\tcc{Online Phase}
$[\delta]_j = ([x]_j - [t]_j) \mod 20$ \;
$\delta = [\delta]_0 + [\delta]_1$ \tcp{reconstruct $\delta$ by $1$ round of communication} 
$p = sin(\delta), q = cos(\delta)$\;
$[y]_j = p[v]_j + q[u]_j$
\end{algorithm}

\section{Fourier Series Fitting Results}\label{sec: fourier fitting}
\begin{figure}[ht]
	\centering
	\includegraphics[scale=0.5]{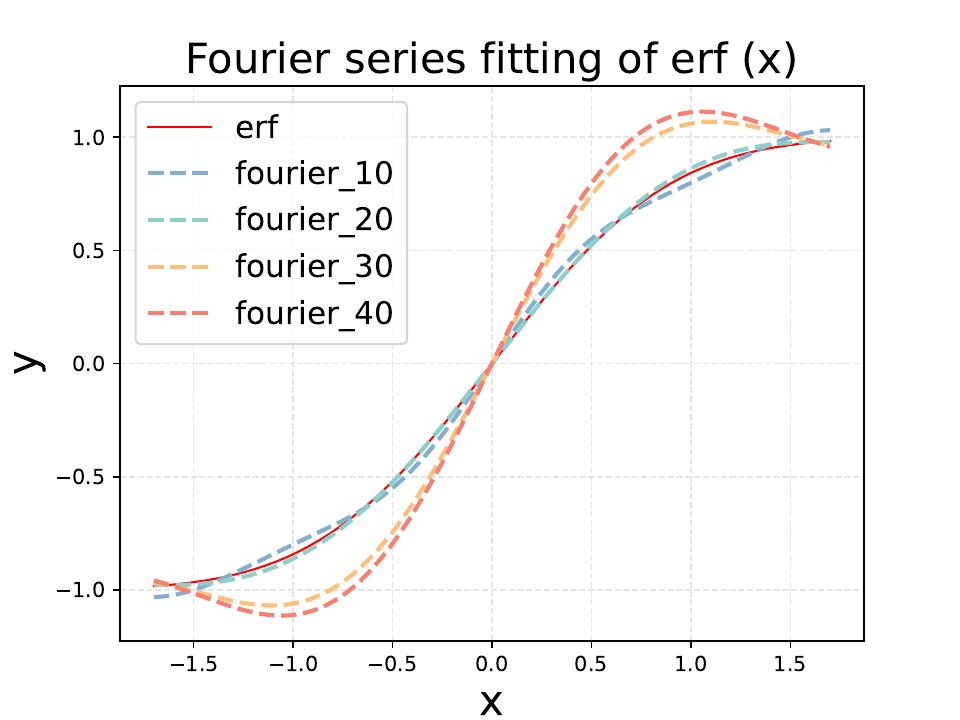} 
	\caption{Fourier series fitting results for different periods. ``$\text{fourier}_{10}$'' denotes that the period of the sin function in the Fourier series is $10$.}
	\label{fig: fourier_fit}
\end{figure}

In this section, we give the results of fitting $\text{erf}(x)$ using Fourier series composed of different periodic sin functions.
Specifically, we fit $\text{erf}(x)$ using the 7-th order Fourier series composed of sin functions with periods of 10, 20, 30, and 40, respectively, and the specific fitting results are shown in \cref{fig: fourier_fit}.

\section{Models and Hyper-parameter}\label{sec: models and hyperparameter}
\paragraph{Models.}
In this section, we provide a concise overview of the architecture of the experimental models. For more detailed information, we refer the readers to the HuggingFace Transformers library.
\begin{itemize}
    \item BERT$_{\text{BASE}}$: BERT$_{\text{BASE}}$ represents the foundational version of the Bert model, comprising 12 Transformer encoder layers, a hidden size of 768, and 12 heads. With 110 million parameters, it undergoes training on a substantial corpus of unlabeled text data.
    \item BERT$_{\text{LARGE}}$: BERT$_{\text{LARGE}}$ serves as an expanded iteration of BERT$_{\text{BASE}}$, featuring 24 Transformer encoder layers, a hidden size of 1024, and 16 heads. Boasting approximately 340 million parameters, this version exhibits increased potency, enabling it to capture intricate language patterns.
\end{itemize}

\paragraph{Hyper-parameter.}
For LayerNorm and Softmax, we set the constants $\eta$ as $2000$ and $5000$, respectively, to ensure that the value of the denominator can be deflated to a reasonable range of convergence. We follow the choice of hyperparameters for fine-tuning and distillation in MPCFormer \citep{li2022mpcformer}. Specifically, in the fine-tuning phase, we use a learning rate of $[1e-6,5e-6,1e-5,1e-4]$, a batch size of $[64,256]$, and epochs of $[10,30,100]$. We fine-tuned each model with a combination of hyperparameters and selected the best performing model as teacher. In the distillation phase, we decide the number of epochs based on the MSE loss of the embedding and transformation layer distillations. For small datasets (CoLA, MRPC, RTE), the batch size is 8; for large datasets (QNLI, STS-B), the batch size is 32.
Specifically, in the embedding and transform layer distillation phases, 10 epochs for QNLI, 20 epochs for MRPC, 50 epochs for STS-B, 50 epochs for CoLA, and 50 epochs for RTE.

\label{sec:appendix}


\end{document}